\documentclass[11pt]{article}
\usepackage{authblk}

\newlength{\defbaselineskip}
\usepackage{graphicx}
\usepackage{amsmath}
\usepackage{amssymb}
\usepackage[caption=false,font=footnotesize]{subfig}
\usepackage{stmaryrd}

\usepackage{algorithmic}
\usepackage{algorithm}

\DeclareMathOperator*{\argmin}{arg\,min}

\begin{document}

\title{
Two-Stage Surrogate Modeling for Data-Driven Design Optimization with Application to Composite Microstructure Generation}

\author{Farhad Pourkamali-Anaraki
  \\
  \textit{Mathematical and Statistical Sciences, University of Colorado Denver, CO, USA}
  \and
  Jamal F. Husseini\\
  \textit{Mechanical and Industrial Engineering, University of Massachusetts Lowell, MA, USA}
  \and 
  Evan J. Pineda\\
   \textit{NASA Glenn Research Center, Cleveland, OH, USA}
   \and 
   Brett A. Bednarcyk\\
   \textit{NASA Glenn Research Center, Cleveland, OH, USA}
   \and 
    Scott E. Stapleton\\
   \textit{Mechanical and Industrial Engineering, University of Massachusetts Lowell, MA, USA}
}

\date{\vspace{-7ex}}

\maketitle

\begin{abstract}
This paper introduces a novel two-stage machine learning-based surrogate modeling framework to address inverse problems in scientific and engineering fields. In the first stage of the proposed framework, a machine learning model termed the ``learner'' identifies a limited set of candidates within the input design space whose predicted outputs closely align with desired outcomes. Subsequently, in the second stage, a separate surrogate model, functioning as an ``evaluator,'' is employed to assess the reduced candidate space generated in the first stage. This evaluation process eliminates inaccurate and uncertain solutions, guided by a user-defined coverage level. The framework's distinctive contribution is the integration of conformal inference, providing a versatile and efficient approach that can be widely applicable. To demonstrate the effectiveness of the proposed framework compared to conventional single-stage inverse problems, we conduct several benchmark tests and investigate an engineering application focused on the micromechanical modeling of fiber-reinforced composites. The results affirm the superiority of our proposed framework, as it consistently produces more reliable solutions. Therefore, the introduced framework offers a unique perspective on fostering interactions between machine learning-based surrogate models in real-world applications.
\end{abstract}

\section{Introduction}\label{sec:into}
Design optimization plays a vital role in engineering and scientific disciplines by identifying configurations of input parameters that produce desired outputs or quantities of interest \cite{kalsi1999comprehensive,padmanabha2021solving,gallet2022structural,liu2022simulation,hariri2022automated}. For example, design optimization in integrated computational materials engineering (ICME) is gaining momentum to find the best candidate in the materials search space that meets specific design criteria rather than relying on inefficient guess-and-check methods \cite{morgan2020opportunities,chen2021polymer,sharma2022advances}. Design optimization is often referred to as an inverse problem because it entails inverting the forward mapping from inputs to outputs while considering application-specific constraints, such as fixing or restricting a portion of the input parameters \cite{noh2020machine}. Inverse problems typically present significant challenges due to complex and nonlinear input-output relationships that may not have analytical representations. Such complexity often precludes the existence of a one-to-one mapping between inputs and outputs, leading to infeasible or multiple possible solutions when solving inverse problems.

An emerging approach to inverse analysis revolves around developing machine learning-based surrogate models. A conventional notion of supervised machine learning is to use observations in the form of input-output pairs, known as the training data set, to learn a proxy or surrogate model that mimics the behavior of complex systems for making predictions about unseen or future inputs \cite{kim2020machine,9328244,campet2020design}. Thus, surrogate models for the forward problem can be used to set up a constrained optimization problem, allowing us to search for inputs that minimize the difference between the target and predicted outputs. Although utilizing machine learning for inverse problems offers significant advantages in applications that require expensive computer simulations \cite{hariri2019matrix}, it is crucial to select appropriate machine learning models and adjust external configuration options, termed hyperparameters. These include choices such as  the regularization strength, the degree parameter in polynomial regression, the number of nearest neighbors, the topology of neural networks, and many others \cite{bischl2021hyperparameter}. Hence, the practical challenge is twofold: first, to find reliable and computationally efficient surrogate models for revealing relationships between inputs and outputs of complex systems \cite{hariri2018support,paleyes2022challenges,POURKAMALIANARAKI2023106983}, and second, to develop rigorous strategies for inspecting solutions of surrogate-assisted inverse problems \cite{singh2021decision}.

To address the issues highlighted, this paper proposes a novel two-stage machine learning-based surrogate modeling framework. In the first stage, a machine learning model, referred to as a \textit{learner}, identifies a small set of candidates in the input design space whose predicted outputs are close to the target output. Thus, the main task in this step is to reduce the size of the search space rather than selecting just the best candidate as the ultimate solution, which is common in existing single-stage inverse analysis techniques. On the other hand, in the second stage of our framework, a separate surrogate model acts as an \textit{evaluator} to audit the reduced space generated in the first stage and eliminate inaccurate solutions based on a user-specified coverage level. Therefore, this framework provides safety measures when there is a discrepancy between the two machine learning-based surrogates.

In our proposed framework, the learner may be any machine learning method and is not limited in form, e.g., we can employ polynomial regression, nearest neighbors regressor, kernel machines \cite{pourkamali2016randomized,pourkamali2020kernel}, Gaussian processes, or neural networks. However, the evaluator model must be designed to quantify the predictive uncertainty associated with each point \cite{einbinder2022training}. To achieve this goal, we propose utilizing conformal inference \cite{lei2018distribution,angelopoulos2023conformal,fontana2023conformal}, a general-purpose statistical technique that allows the determination of prediction intervals and corresponding coverage levels for each candidate in the design space. This approach provides lower and upper limits of the prediction interval for each candidate solution, allowing us to discard those candidates whose constructed intervals do not include the target value of the design optimization problem. As a result, the proposed framework can be viewed as a form of data-driven regularization to prevent the inverse problem from relying too heavily on the performance of the learner model.

\begin{figure}[t]
\centering
\includegraphics[width=0.98\linewidth]{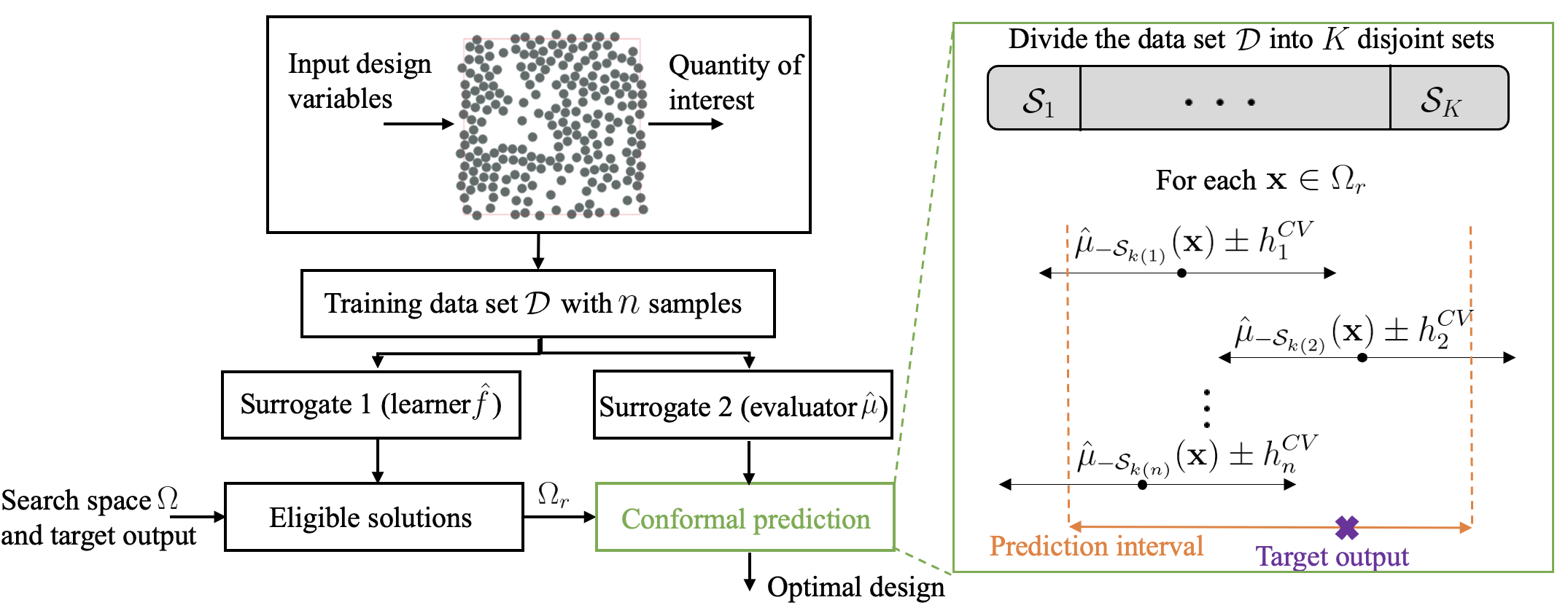}
\caption{\label{fig:overall}
Illustrating the proposed data-driven design optimization framework using two surrogate models, which we refer to them as learner and evaluator models. The learner model $\hat{f}$ serves as a proxy for the forward model to narrow down the original search space by finding a small subset of eligible solutions $\Omega_r\subset\Omega$. The evaluator agent $\hat{\mu}$, which is distinct from the learner, uses conformal prediction to obtain prediction intervals for each member of the reduced search space $\Omega_r$ to verify that the target output falls within the constructed interval. Hence, this framework provides safety measures when there is a discrepancy between the two machine learning-based surrogates.  
}
\end{figure}

The proposed two-stage surrogate modeling framework for data-driven design optimization, depicted in Figure \ref{fig:overall}, offers a high degree of flexibility and ease-of-use because of reducing the need for extensive hyperparameter tuning and human oversight. Instead of relying on a single surrogate model, we allow two distinct surrogate models to interact with each other to identify discrepancies. Additionally, the framework is not limited to machine learning algorithms that inherently provide uncertainty estimates, such as Gaussian processes \cite{schulz2018tutorial} and Bayesian methods \cite{izmailov2021bayesian,guo2022bayesian}, because the second stage relies on conformal prediction. In fact, any standard machine learning algorithm that produces point estimates can serve as the evaluator model, followed by a statistical procedure to account for the variability in the fitted regression model. Another advantage of the proposed framework is that it can seamlessly handle multiple quantities of interest by verifying that each target value falls into the corresponding prediction interval. 

To the best of our knowledge, this paper presents the first application of conformal prediction for the inverse analysis of engineering systems. Hence, comprehensive numerical investigations are presented to demonstrate the significance of the two-stage surrogate modeling approach to solve inverse problems and remove inconsistent solutions. To this end, this paper considers a benchmark problem widely used for uncertainty quantification and sensitivity analysis, called the Ishigami function, to solve design optimization problems with one and two desired quantities of interest. In addition, this paper focuses on an engineering application concerning micromechanical modeling of fiber reinforced composites. In this application, the input design space captures the spatial variability of fiber arrangements and the goal is to generate microstructures with desired descriptors. Typical microstructure generators for fiber reinforced composites are hard-core generators which rely on guess-and-check methods for placing fibers. Therefore, surrogate-assisted inverse analysis can substantially accelerate the optimization process in this application.

While this paper focuses on the application of the two-stage surrogate modeling framework in mechanical engineering, it is important to recognize related inverse problems highlighted in the recent literature. For example, previous studies by El Sayed et al.~\cite{el2021novel,el2022novel} focused on modeling supply chain networks and transportation problems. These efforts involved identifying ideal solutions, followed by using a different optimization problem to further enhance and refine these solutions. Furthermore, Zhuang et al.~\cite{9979725} showed how to sequentially adjust the control input, with the objective of reducing errors and improving the overall performance of the system. In addition, Djordjevic et al.~\cite{djordjevic2023data} introduced adaptive dynamic programming (ADP) to design closed-loop control systems, which was applied to a linear model of hydraulic servo actuators.

Finally, we discuss connections between the proposed two-stage surrogate modeling framework and existing metaheuristic algorithms \cite{agrawal2021metaheuristic,zhao2022dandelion,gholian2023metaheuristic}. In these types of algorithms, the main goal is to explore the solution space to discover near-optimal solutions without relying on a specific problem's structure. To explain the basic idea, a learner identifies promising regions, while a distinct evaluator model assesses the quality of these solutions to enhance them. What sets our work apart is the novel utilization of conformal inference to offer feedback on the predictive uncertainty of the learner model. Therefore, this integration of conformal prediction methods provides a unique perspective to enable diverse machine learning models to interact with each other to remove uncertain/inaccurate solutions of design optimization problems, as validated by a complex engineering application. 

This paper is structured as follows. Section \ref{sec:prelim} introduces the notation and provides a brief mathematical formulation of design optimization and machine learning problems and their interrelationship. In Section \ref{sec:proposed}, we propose a two-stage machine learning-based surrogate modeling framework for solving design optimization problems, including the use of conformal prediction. To demonstrate the effectiveness of our proposed approach for nonlinear systems in controlled environments with one and two target outputs, we present numerical experiments using a benchmark problem in Section \ref{sec:bench}. Then, in Section \ref{sec:mic}, we focus on the composite microstructure generation problem and provide a thorough examination of design optimization for finding optimal fiber arrangements to meet various design criteria. Finally, in Section \ref{sec:conc}, we conclude by presenting remarks on the merits and limitations of the current work, as well as discussing future research directions.

\section{Preliminaries and Notations}\label{sec:prelim}
Let us consider a system with $p$ input design variables in the form of a vector $\mathbf{x}=[x_1,\ldots,x_p]\in\mathbb{R}^p$ and a response variable or output $y\in\mathbb{R}$. Throughout this paper, we make the assumption that boldface letters represent vectors to differentiate them from scalar quantities. Moreover, for simplicity, we study systems with one quantity of interest in this section, and we will discuss the extension to multiple outputs in the next section.  Given the forward model $f$ that explains the underlying behavior of this system, we can formulate the \textit{observation model} as $y=f(\mathbf{x})+\delta y$, where $\delta y$ represents a noise term accounting for possible observation errors and uncertainty sources. For example, $\delta y$ can arise from the aleatoric uncertainty reflecting the inherent randomness in many engineering applications \cite{hariri2022structural}. In Section \ref{sec:mic}, we will illustrate the role of the noise term $\delta y$ in the context of the microstructure generation problem.

Since the forward model $f$ is typically complex and does not have a closed-form expression, machine learning (ML) methods provide valuable tools to construct an approximate yet tractable proxy or surrogate model \cite{cai2021surrogate,lei2021bayesian,stuckner2021optimal,hearley2023robust,nasrin2023active}. To this end, the first step is to collect $n$ observations in the form of input-output pairs, i.e., $\mathcal{D}=\{(\mathbf{x}^{(1)}, y^{(1)}), \ldots,(\mathbf{x}^{(n)}, y^{(n)}) \}$, which is known as the training data set \cite{nasrin2023active}. Then, an ML learner configured by hyperparameters $\lambda$ maps the training data set $\mathcal{D}$ to the surrogate model $\hat{f}$, i.e., $(\mathcal{D},\lambda)\mapsto\hat{f}$. Hyperparameters are external parameters of ML algorithms that remain fixed during the training stage and their values must be provided in advance. For example, in the case of polynomial regression, the polynomial degree is a crucial hyperparameter that should be adjusted ahead of time to control the complexity of the surrogate model. 

After setting hyperparameters, the training stage involves finding internal parameters or weights, e.g., coefficients of polynomial terms, using the concept of empirical risk minimization \cite{murphy2022probabilistic}:
\begin{equation}
r^{\text{emp}}(\tilde{f}):=\frac{1}{|\mathcal{D}|}\sum_{(\mathbf{x}^{(i)}, y^{{(i)}})\in\mathcal{D}} l\big(y^{(i)}, \tilde{f}(\mathbf{x}^{(i)})\big),\;\;\hat{f}=\argmin_{\tilde{f}} r^{\text{emp}}(\tilde{f}),\label{eq:ERM}
\end{equation}
where $|\mathcal{D}|$ denotes the number of data points in $\mathcal{D}$. The loss function $l$ measures the difference between the actual and predicted outputs. For example, the quadratic loss function $l\big(y^{(i)}, \tilde{f}(\mathbf{x}^{(i)})\big)=(y^{(i)} - \tilde{f}(\mathbf{x}^{(i)}))^2$ is popular for regression tasks and the objective function in Eq.~\eqref{eq:ERM} is known as the mean squared error in this case.
In the subsequent sections, we often use the subscript $\mathcal{D}$ to emphasize $\hat{f}_{\mathcal{D}}$ is fitted to the data set $\mathcal{D}$ because in some cases we will use a subset of $\mathcal{D}$ to form the empirical risk minimization problem for model fitting.

Using the surrogate model $\hat{f}$ approximating the underlying forward model $f$, inverse analysis can be written as a constrained optimization problem to search for the best candidate given the target output $y^{\text{target}}$ and the search space $\Omega\subset \mathbb{R}^p$:
\begin{equation}
\argmin_{\mathbf{x}\in\Omega} \big(\hat{f}(\mathbf{x})- y^{\text{target}}\big)^2+\gamma r^{\text{reg}}(\mathbf{x}).\label{eq:search}
\end{equation}
The first term measures the misfit between the target output and those predicted by the surrogate model $\hat{f}$, whereas the second term is the so-called regularization term which does not depend on $\hat{f}$. Regularization terms can be used to introduce additional constraints on the optimization problem. This is particularly valuable when dealing with design specifications, limits, or constraints that need to be considered during the optimization process.
Here, $\gamma\geq 0$ denotes the regularization strength and choosing $\gamma=0$ implies that the solution of the inverse problem solely relies on the predictive behavior of $\hat{f}$. Hence, the solution of the above optimization problem is susceptible to the error term $\delta y$ in the observation model that was incurred when collecting the training data set $\mathcal{D}$. Also, finding the right hyperparameter values is critical because the surrogate model $\hat{f}$ must provide accurate predictions across the entire design space. 

However, as we increase the amount of regularization $\gamma$, using the regularization function $r^{\text{reg}}$ allows us to alleviate dependencies on the surrogate model $\hat{f}$ and overcome the problems of non-uniqueness and noisy observations by introducing prior knowledge. For example, one popular choice of regularization is known as the Tikhonov regularization \cite{ghattas2021learning}, where $r^{\text{reg}}(\mathbf{x})=\|\mathbf{x}\|_2^2=x_1^2+\ldots+x_P^2$. Additionally, recent works have incorporated detailed domain expertise into designing application-specific and physics-informed regularization functions \cite{puel2022mixed,jagtap2022physics,weissmann2022adaptive}. 

The search space $\Omega$ for the optimization problem in Eq.~\eqref{eq:search} is the set of inputs in the design space $\mathbb{R}^p$, where the entries in each dimension are either fixed or distributed between predefined minimum and maximum values. Unlike the training data set $\mathcal{D}$, the search space is unlabeled because of lacking ground truth outputs and high computational costs to find $f(\mathbf{x})$ for all inputs in $\Omega$. Formally, we define the set $\Omega:=\{\mathbf{x}\in\mathbb{R}^p: g_1(\mathbf{x})\leq 0,\ldots, g_{n_1}(\mathbf{x})\leq 0, s_1(\mathbf{x})=0,\ldots,s_{n_2}(\mathbf{x})=0\}$, where $g_i$ and $s_i$ denote application-specific inequality and equality constraints, respectively. For example, an equality constraint $s_i(\mathbf{x})=x_i-x_i^*=0$ means that the value of the $i$-th design parameter is fixed and must be equal to $x_i^*$ during the optimization process. Similarly, an inequality constraint $g_i(\mathbf{x})=x_i-x_i^u\leq0$ means that the value of $x_i$ must not exceed the designated upper bound $x_i^u$. This paper focuses on using uniform sampling to discretize the given interval for each adjustable input parameter so that $\Omega$ has a finite number of elements. Thus, we can compute the value of the objective function in Eq.~\eqref{eq:search} for each $\mathbf{x}\in\Omega$ to find the optimal input that minimizes the objective function.  

While the single-stage surrogate modeling approach for solving design optimization problems presented in Eq.~\eqref{eq:search} can be effective in certain cases, it has two critical shortcomings that should be considered. First, adopting ML methods in new applications to find the surrogate model $\hat{f}$ requires significant computational resources and expertise because well-performing surrogates are not immediately available and selecting the wrong model can lead to inaccurate or unreliable results. Hence, the user must perform an extensive grid search over a wide range of hyperparameter configurations or employ advanced hyperparameter optimization (HPO) methods \cite{bischl2023hyperparameter} to identify the top-performing surrogate model. Despite these efforts,  the sensitivity of the resulting surrogate model concerning the noise term $\delta y$ in the observation model and the selected hyperparameter values is typically unknown, thus the inverse problem may fail without warning. Consequently, the inverse analysis quality strongly depends on the mapping $(\mathcal{D}, \lambda)\mapsto \hat{f}$ and the choice of $\mathcal{D}$ and $\lambda$. The second challenge is related to designing appropriate regularization functions $r^{\text{reg}}$ and adjusting the regularization strength $\gamma$, which demands detailed domain expertise and numerous trial-and-error cycles to understand the impact of regularization on the accuracy of the inverse problem.

\begin{figure}[ht]
\centering
\includegraphics[width=0.95\linewidth]{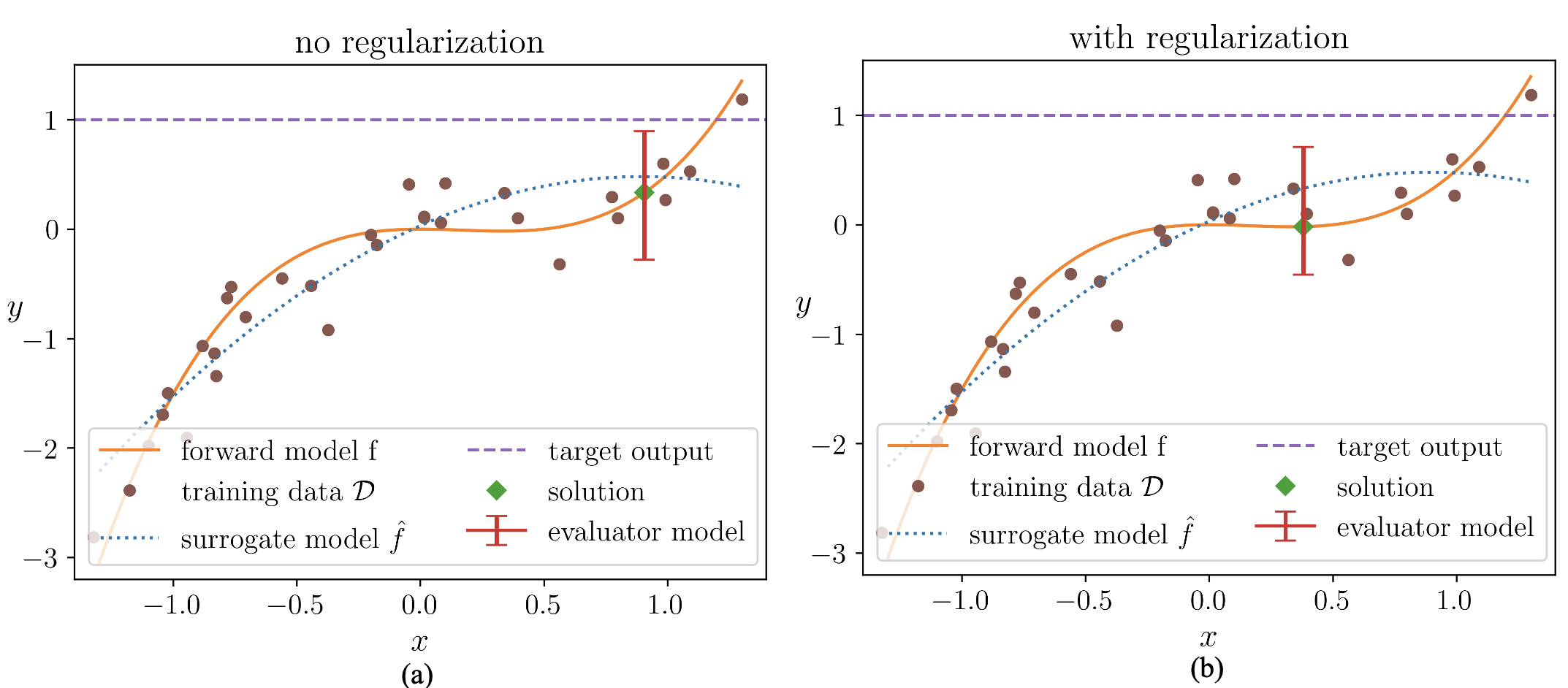}
\caption{\label{fig:syn} Illustrating the shortcomings of single-stage surrogate-assisted inverse problems using a synthetic data set with two values of regularization strength: (a) $\gamma=0$ and (b) $\gamma=1$. This experiment suggests that the optimization problem is oblivious to possible inaccuracies of the surrogate model $\hat{f}$. Thus, there needs to be an effective strategy in place to reject inconsistent solutions, such as using an evaluator model to create a prediction interval, which is shown using a red vertical line. Here, both constructed prediction intervals do not contain the target output. 
}
\end{figure}

To exemplify these shortcomings in a controlled environment, we consider a data set with one input variable $x\in\mathbb{R}$ and the quantity of interest or output takes the form of $y=x^3-0.5x^2+\delta y$, where $\delta y$ is chosen to be a zero-mean normal random variable. As shown in Figure \ref{fig:syn}, we collect a data set $\mathcal{D}$ containing $n=30$ observations to train the surrogate model $\hat{f}$, which is assumed to be a degree-$2$ polynomial function, thus creating a minor model misspecification. The polynomial degree is a critical hyperparameter that must be set before starting the learning process. While Figure \ref{fig:syn}(a) shows that $\hat{f}$ provides a reasonable approximation of the underlying input-output mapping, it is apparent that solving the inverse problem for the target output value of $1$ without any regularization leads to an inaccurate solution for which the output value is closer to $0$ rather than $1$. 

We also observe that adding the Tikhonov regularization with $\gamma=1$ in Figure \ref{fig:syn}(b) moves the solution to the left, which is undesirable. Hence, this experiment exemplifies the need for rigorous strategies to inspect the solution of single-surrogate inverse problems in decision-critical engineering applications. This motivates us to design a flexible data-driven regularization technique in the next section by introducing an evaluator model to provide prediction intervals for each possible solution and verify that the target output falls into the constructed interval. Building such intervals is necessary for problems with continuous outcomes because the probability of a continuous random variable taking a particular value is zero. As depicted in Figure \ref{fig:syn}, the evaluator model in this case study, which is a $4$-nearest neighbor regressor, can pinpoint that the solution of the optimization problem is invalid because the target output value is outside the prediction interval shown in red. 

\section{Proposed Two-Stage Surrogate Modeling Framework}\label{sec:proposed}
In this section, we present an enhanced approach for regularizing and inspecting inverse problems using a new two-stage surrogate modeling framework. The main motivation is to identify a suitable regularizer directly from the data rather than relying on domain expertise to design the regularization term. Similar to existing surrogate-assisted inverse problems explained in the previous section, the first stage involves training a regression model $\hat{f}_{\mathcal{D}}$ that serves as a substitute for the unknown forward model. However, instead of identifying a single model input that most nearly matches the target output, or the minimizer of the objective function in Eq.~\eqref{eq:search}, our goal in the first stage of the proposed framework is to determine a predefined number of eligible solutions and eliminate the need for meticulously tailored regularization functions.

To achieve this goal, we assess the suitability of all inputs in the search space $\Omega$ by computing the squared difference between the predicted and target output without utilizing any regularization term:
\begin{equation}
e(\mathbf{x}):=(\hat{f}_{\mathcal{D}}(\mathbf{x}) - y^{\text{target}})^2,\;\mathbf{x}\in\Omega.\label{eq:dist}
\end{equation}
This step is computationally affordable because  $\Omega$ is a finite set and function evaluations are inexpensive once a surrogate model $\hat{f}_{\mathcal{D}}$ is trained. Subsequently, we identify the $B$ smallest values of $e$ along with their corresponding inputs from the search space $\Omega$. It is worth noting that when $B=1$, this stage is equivalent to finding the optimal solution that minimizes the objective function in Eq.~\eqref{eq:search}. However, due to the presence of nonlinear relationships in complex engineering systems, it is plausible that multiple solutions may exist and we consider $B>1$ for this reason. Consequently, the first stage of our proposed framework enables us to discover the top $B$ eligible solutions, thereby constructing the reduced set $\Omega_r \subset \Omega$ that contains these inputs within the design space. This strategy facilitates a more comprehensive exploration of potential solutions, acknowledging the possibility of multiple viable outcomes within the given problem context. 

With this motivation in place, the second stage of the proposed framework integrates conformal prediction with a distinct ML model that we call an evaluator model to construct prediction intervals. Let $\hat{\mu}$ and $1-\alpha$ represent the evaluator model and a target coverage level specified by the user, respectively. The key idea behind this stage is to construct a \textit{prediction interval} $\hat{C}_{\alpha}$ that satisfies the following probabilistic argument for each $\mathbf{x}\in\Omega_r$ \cite{tibshirani2019conformal}:
\begin{equation}
\text{Prob}\big(y\in\hat{C}_{\alpha}(\mathbf{x})\big)\geq 1-\alpha,
\end{equation}
where $y$ denotes the ground truth output for the input data point $\mathbf{x}$. Recall that $y$ is unknown because the search space is unlabeled. Therefore, the above argument tells us the probability that the actual output of $\mathbf{x}$ lies in the interval $\hat{C}_{\alpha}(\mathbf{x})$ should be greater than or equal to the target level $1-\alpha$. Conversely, the probability that the ground truth output does not belong to this interval would be lower than $\alpha$ or the miscoverage level. Thus, $\alpha$ should be reasonably small in conformal prediction and it is common to choose $\alpha\in[0.05,0.5]$. 

While there are several forms of conformal prediction, we use a recently modified version that builds on the concept of $K$-fold cross-validation (CV) \cite{bates2023cross} to obtain a delicate trade-off between accuracy and efficiency. Suppose that we divide the training data set $\mathcal{D}$ into $K$ disjoint subsets $\mathcal{S}_1,\ldots,\mathcal{S}_K$ of equal sizes. Without loss of generality, we assume $n/K$ is an integer. We define $\hat{\mu}_{-\mathcal{S}_k}$ to be the fitted evaluator regression model when the $k$-th subset removed, i.e., the evaluator model $\hat{\mu}$ is fitted to $\mathcal{D}\setminus \mathcal{S}_k$. Moreover, let $k(i)\in\{1,\ldots,K\}$ determine the subset that contains the $i$-th training sample $(\mathbf{x}^{(i)},y^{(i)})$, $i=1,\ldots,n$. Then, the lower and upper limits of the prediction interval can be found as: 
\begin{equation}
\hat{C}_{\alpha}(\mathbf{x})=\Big[ q_{\alpha}^-\big\{ \hat{\mu}_{-\mathcal{S}_{k(i)}}(\mathbf{x})-h_i^{CV}\big\}, q_{\alpha}^+\big\{ \hat{\mu}_{-\mathcal{S}_{k(i)}}(\mathbf{x})+h_i^{CV}\big\}\Big],\label{eq:interval}
\end{equation}
where $h^{CV}_i$ is defined as the holdout residual for the training sample with the index $i$:
\begin{equation}
h^{CV}_i:=\Big|y^{(i)}-\hat{\mu}_{-\mathcal{S}_{k(i)}}(\mathbf{x}^{(i)})\Big|,\;i=1,\ldots,n.
\end{equation}
Also, $q_{\alpha}^+$ denotes the $(1-\alpha)$ quantile of the $n$ calculated values in Eq.~\eqref{eq:interval} and $q_{\alpha}^-$ is defined as the $\alpha$ quantile \cite{barber2023conformal}. This process of constructing prediction intervals is illustrated in right panel of Figure \ref{fig:overall}, where we find a prediction interval for an eligible solution of the first stage. 

From the computational standpoint, this variant of conformal prediction allows us to control the number of times we must train the evaluator model $\hat{\mu}$, which equals the number of folds $K$ in CV. Thus, in practice, we often choose $K \ll n$ in order to avoid incurring excessive computational costs. Furthermore, another computational benefit of conformal prediction is that finding the residual term $h^{CV}_i$ does not rely on the choice of $\Omega_r$, so we have to compute $h^{CV}_1,\ldots,h^{CV}_n$ just once for all eligible solutions, leading to substantial savings for large values of  $B$ in the first stage.   

After using conformal prediction to obtain the prediction interval for each $\mathbf{x}\in\Omega_r$, we perform a simple inspection step to ensure that the target output of the inverse problem, denoted as $y^{\text{target}}$, falls within the constructed interval $\hat{C}_{\alpha}(\mathbf{x})$. If the target output is not contained within this interval, i.e., $y^{\text{target}}\notin \hat{C}_{\alpha}(\mathbf{x})$, it indicates a discrepancy between the learner and evaluator, rendering the solution $\mathbf{x}$ non-viable. Consequently, we begin by sorting the eligible solutions of the first stage in ascending order based on their distances to the target output predicted by the learner model. We then examine these potential solutions in the given order until we find a solution that aligns with the evaluator model, confirming that the target output lies within the interval predicted by the evaluator model. Hence, we can combine conformal prediction with any ML model to produce intervals that contain the ground truth output with user-specified coverage levels. 

In the last part of this section, we expand the proposed two-stage surrogate modeling framework to systems that involve multiple quantities of interest. Suppose that the system has $t$ outputs, and the target output of the inverse problem can be represented as a vector $\mathbf{y}^{\text{target}}$ in $\mathbb{R}^t$. Recall that we use boldface letters for representing vectors. The learner model, which serves as a substitute for the forward model, is also a multi-output function denoted as $\hat{f}_{\mathcal{D}}: \mathbb{R}^p\rightarrow\mathbb{R}^t$. Consequently, the process of identifying eligible solutions $\Omega_r$ within the search space can be modified as follows:
\begin{equation}
e(\mathbf{x})=\|\hat{f}_{\mathcal{D}}(\mathbf{x}) - \mathbf{y}^{\text{target}}\|_2^2,\;\mathbf{x}\in\Omega.\label{eq:dist2}
\end{equation}
Furthermore, assume that the evaluator model takes the form of $\hat{\mu}:
\mathbb{R}^p\rightarrow \mathbb{R}^t$, thus the holdout residual for the training sample $(\mathbf{x}^{(i)},\mathbf{y}^{(i)})$ can be written as: 
\begin{equation}
\mathbf{h}^{CV}_i:=\Big|\mathbf{y}^{(i)}-\hat{\mu}_{-\mathcal{S}_{k(i)}}(\mathbf{x}^{(i)})\Big|\in\mathbb{R}^t,\;i=1,\ldots,n,
\end{equation}
where we take the element-wise absolute value of the difference between the ground truth and predicted output vector. As a result, we can find the lower and upper limits of the prediction interval for each quantity of interest, indexed by $j\in\{1,\ldots,t\}$, using the two quantiles across the $j$-th entry of $q_{\alpha}^-\{\hat{\mu}_{-\mathcal{S}_{k(i)}}(\mathbf{x})-\mathbf{h}_i^{CV}\}$ and $q_{\alpha}^+\{\hat{\mu}_{-\mathcal{S}_{k(i)}}(\mathbf{x})+\mathbf{h}_i^{CV}\}$, $i=1,\ldots,n$. In this case, the final step to find a viable solution involves verifying that all the $t$ entries of $\mathbf{y}^{\text{target}}$ fall into the corresponding prediction intervals found by the evaluator model $\hat{\mu}$.

\section{Benchmark Problem}\label{sec:bench}
In this section, we utilize the Ishigami function \cite{allaire2012variance,hariri2022structural} as a benchmark to showcase the performance and user-friendly nature of our proposed two-stage surrogate modeling framework. The choice of the Ishigami function as our benchmark is motivated by the absence of any suitable regularization function for the inverse problem at hand. Consequently, we can effectively highlight the advantages of employing a distinct evaluator model to regulate the design optimization problem in a data-driven manner, eliminating the the need for prior knowledge and resource-intensive search cycles. The Ishigami function is defined as follows for a $3$-dimensional input vector $\mathbf{x}=[x_1,x_2,x_3]$:
\begin{equation}
f(\mathbf{x})=\sin(x_1)+a\sin(x_2)^2+b x_3^4\sin(x_1). \label{eq:ishigami}
\end{equation}
In this function, $x_1, x_2, x_3$ are independent variables that are uniformly distributed in the interval $[-\pi,\pi]$, while $a$ and $b$ are two constants. This particular function possesses two challenging properties that make it suitable for capturing subtle relationships within engineering systems. Firstly, it incorporates nonlinear terms like sine and power functions. Secondly, the Ishigami function is not monotonic, meaning it does not consistently increase or decrease with respect to the three independent variables. These properties add complexity to the function, making it an excellent benchmark for evaluating the performance of our methods.

\begin{figure}[ht!]
	\centering
	\includegraphics[width=0.95\linewidth]{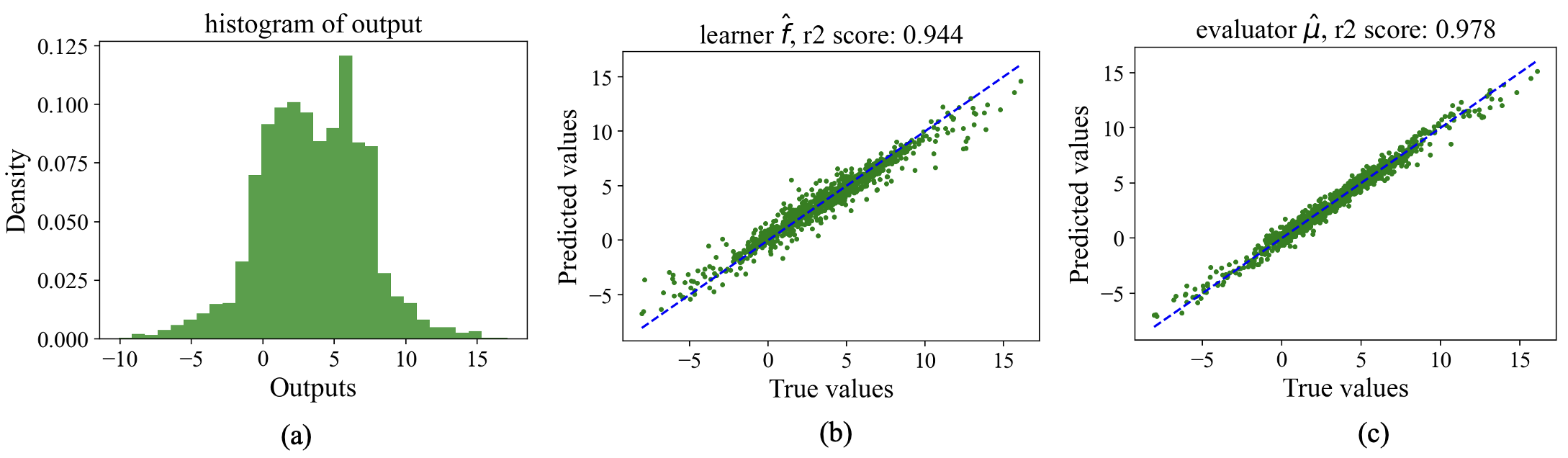}
	\caption{\label{fig:bench-visual}
		Using the Ishigami function to generate the training data set $\mathcal{D}$ and fitted surrogate models $\hat{f}$ and $\hat{\mu}$. The learner is a nearest neighbor regressor, while we adopt a polynomial regression model as an evaluator. To measure the predictive performance of these models, we report predicted vs.~true outputs on a separate test data set that was not used for training.
	}
\end{figure}

To start our analysis, we focus on inverse problems that revolve around a single target output $y^{\text{target}}$. In this scenario, we keep the input parameter $x_1$ fixed and aim to determine the optimal configuration of the remaining two inputs, $x_2$ and $x_3$, that yields the desired target output.
To establish our observation model, we set the values of $a$ and $b$ to be $7$ and $0.1$, respectively. We then gather a total of $n=2,\!000$ training data points, which will be employed to construct our two surrogate models $\hat{f}$ and $\hat{\mu}$. The distribution of outputs in the collected training data set $\mathcal{D}$ is shown in Figure \ref{fig:bench-visual}(a). The learner model $\hat{f}$ is set to be a nearest neighbor regressor, where we conduct a grid search to adjust the number of nearest neighbors that reveals the optimal number of nearest neighbors is $6$. We proceed to visualize the predicted outputs against the actual outputs using a separate test data set comprising $1,\!000$ samples in Figure \ref{fig:bench-visual}(b). Furthermore, we report the coefficient of determination, often denoted as R-squared or r2, for the trained regression model, which serves as a measure of how effectively unseen samples can be predicted by the model. In a formal way, an r2 value of 1 indicates that the model perfectly predicts the dependent variable, while a value of 0 means that the model does not explain any of the variability \cite{nasrinpolymer}. The obtained r2 score is $0.944$, indicative of the model's strong predictive capability, albeit slightly below the best possible score of $1$. To ensure a fair comparison, we employ this surrogate model $\hat{f}$ for both single-stage and our proposed two-stage inverse problems.

As previously mentioned, a key requirement in our framework is that the evaluator model $\hat{\mu}$ must be distinct from the learner model. To this end, we employ a polynomial regression model as our evaluator, and after conducting a grid search, determine that the optimal degree for the polynomial is 6. We utilize the evaluator model $\hat{\mu}$ to plot the predicted outputs versus the ground truth outputs in Figure \ref{fig:bench-visual}(c). Evaluating the r2 score reveals that the evaluator model slightly outperforms the learner model. It is worth noting that the two surrogate models exhibit different predictive behaviors across the entire input space, as indicated by the scatter plot being more concentrated around the 45-degree line for the polynomial regression model. This discrepancy in predictive behavior highlights the importance of considering the interactions between the learner and evaluator models to solve inverse problems. 

\begin{figure}[ht!]
	\centering
	\includegraphics[width=0.9\linewidth]{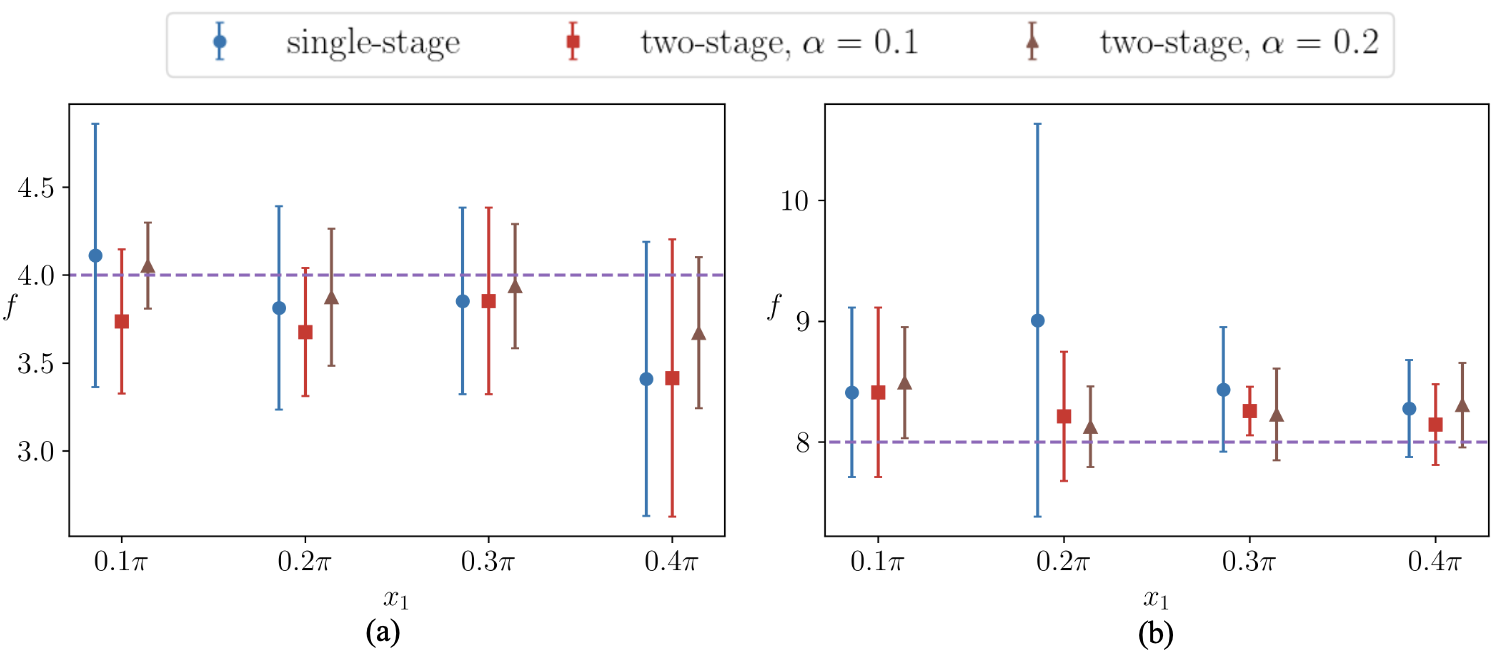}
	\caption{\label{fig:bench-one}
		Solving the inverse problem using the Ishigami function with a scalar target output, shown as a horizontal line, and fixed value of $x_1\in\{0.1\pi,0.2\pi,0.3\pi,0.4\pi\}$. The goal is to find the optimal configuration of $x_2$ and $x_3$ that produces the target output. We plot the mean and standard deviation of actual outputs for the solutions of the inverse problem across $20$ independent trials to capture the randomness associated with the search space $\Omega$. The proposed two-stage surrogate modeling framework is effective in terms of reducing the standard deviation and moving the average output value closer to the target output of the inverse problem.
	}
\end{figure}

The search space $\Omega$ for the inverse problem consists of a fixed value for $x_1$, and it encompasses $10,\!000$ combinations of the remaining two inputs, which are sampled uniformly at random from the two-dimensional space $[-\pi,\pi]^2$.
In the first stage of our proposed framework, we choose $B=10$ eligible solutions to significantly reduce the size of the original search space by a factor of $1,\!000$. Also, we consider two different values of $\alpha$, namely $\alpha=0.1$ and $\alpha=0.2$, for the conformal prediction method utilized in the second stage. Figure \ref{fig:bench-one} presents a comparison of the solutions obtained for the inverse problems under two different target output values  (a) $y^{\text{target}}=4$ and (b) $y^{\text{target}}=8$. 
To evaluate the accuracy of the solutions, we employ the Ishigami function in Eq.~\eqref{eq:ishigami} to determine the actual output for each solution. Moreover, we perform $20$ independent trials to account for the randomness within the search space $\Omega$, thus reporting the mean and standard deviation of the ground truth outputs. This allows us to explore the sensitivity of the obtained solutions concerning the randomness associated with the search space $\Omega$.

Based on the reported results in Figure \ref{fig:bench-one}, it is evident that while the learner model $\hat{f}$ performs well in predicting outputs for unseen test data points, it struggles to accurately solve the single-stage inverse problem. This is exemplified in Figure \ref{fig:bench-one}(b), where we observe a significant deviation from the target output value of $8$, with an average output of approximately $9$ when $x_1=0.2\pi$. Fortunately, we see that the proposed two-stage surrogate modeling framework can greatly enhance the quality of the inverse problem.
For instance, in the case of $x_1=0.2\pi$, the two-stage framework yields a remarkable improvement, as the average target value approaches $8$. At the same time, the standard deviation is reduced by almost a factor of $5$ compared to the single-stage counterpart. Overall, the proposed framework provides more reliable solutions across different values of $x_1$ in Figure \ref{fig:bench-one}(b). 

Moreover, in Figure \ref{fig:bench-one}(a), we observe that the maximum absolute value of the difference between the mean value of our framework with $\alpha=0.2$ and the target output is approximately $0.32$, which is less than $10\%$ of the original target output. Consequently, the evaluator model $\hat{\mu}$ effectively identifies and rejects inaccurate solutions. It is worth noting that the choice of $\alpha$ in conformal prediction, the measure of confidence in the constructed intervals, does not significantly impact the ultimate quality of the inverse problem. Both values of $\alpha$ perform well. Hence, this experiment exhibits that constructing prediction intervals using the probabilistic argument $\text{Prob}(y\in\hat{C}_{\alpha}(\mathbf{x}))\geq 1-\alpha$, $\alpha\in\{0.1,0.2\}$, works well across different values of $x_1$ and $y^{\text{target}}$.  

In the last experiment of this section, we evaluate the performance of the proposed framework for solving inverse problems with more than one target output. To this end, we consider an extension of the Ishigami function to produce two outputs as follows: 
\begin{equation}
f(\mathbf{x})=\begin{bmatrix} f_1 \\ f_2\end{bmatrix}=\begin{bmatrix} \sin(x_1)+7\sin(x_2)^2+0.1x_3^4\sin(x_1) \\ 0.1\big(\sin(x_1)+7\sin(x_2)^2+0.05 x_3^4\sin(x_1)\big)\end{bmatrix}.\label{eq:2d}
\end{equation}
Hence, similar to the previous case study, we fix the value of $x_1$ to search for the best configuration of $x_2$ and $x_3$ to produce the desired target output vector $\mathbf{y}^{\text{target}}\in\mathbb{R}^2$. Hence, we collect the data set $\mathcal{D}$ of size $n=2,\!000$ to train the learner and evaluator models $\hat{f}$ and $\hat{\mu}$, respectively. 

\begin{figure}[ht!]
	\centering
	\includegraphics[width=0.9\linewidth]{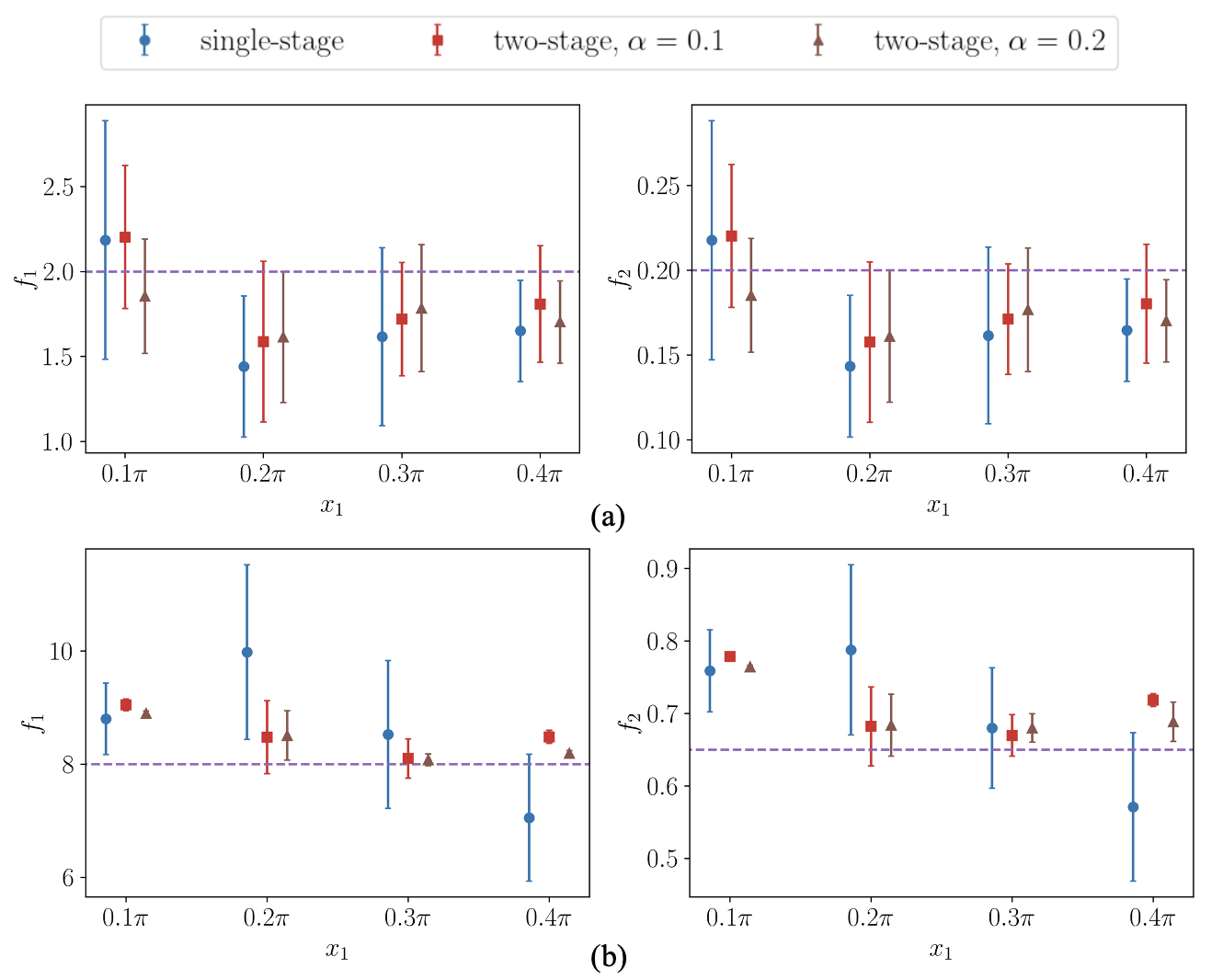}
	\caption{\label{fig:bench-two}
		Solving the design optimization problem using the multi-output Ishigami function with a target output vector and fixed value of $x_1\in\{0.1\pi,0.2\pi,0.3\pi,0.4\pi\}$. The goal is to find the optimal configuration of $x_2$ and $x_3$ that produces the two target outputs $f_1$ and $f_2$. We plot the mean and standard deviation of actual outputs for the solutions of the inverse problem across $20$ independent trials to capture the randomness associated with the search space $\Omega$. 
	}
\end{figure}

Figure \ref{fig:bench-two} presents a comparison of results for two different realizations of the target vector $\mathbf{y}^{\text{target}}$. In Figure \ref{fig:bench-two}(a), the target outputs are set to be $2$ and $0.20$, respectively. Also, the target outputs are $8$ and $0.65$ in Figure \ref{fig:bench-two}(b). In both cases, we deliberately set the target value for $f_2$ to be smaller than that of $f_1$ because we multiplied the second dimension of the Ishigami function by $0.1$ in Eq.~\eqref{eq:2d}. Once again, these results underscore the limitation of single-stage inverse problems since they tend to exhibit high standard deviations and the mean values are often too far from the corresponding target output, which is undesirable. On the other hand, the proposed framework effectively mitigates this issue by reducing the standard deviation across various choices of the search space and $\alpha$. Moreover, on average, the solutions obtained by our framework approach the actual target output more closely. For example, the mean value obtained by our proposed framework with $\alpha=0.2$ is approximately $0.68$ for $x_1\in\{0.2\pi,0.3\pi, 0.4\pi\}$, which is close to the target output of $0.65$ on the right panel of Figure \ref{fig:bench-two}(b).

Overall, these findings demonstrate the significant advantage of the proposed framework over single-stage approaches, as it addresses the shortcomings associated with unsupervised optimization in inverse problems. Therefore, the proposed framework can be viewed as a data-driven regularization technique to inspect the surrogate model used as a proxy model for the forward problem. Furthermore, we highlight the seamless integration of conformal inference with various machine learning models and multiple quantities of interest.

\section{Composite Microstructure Generation}\label{sec:mic}

Fiber reinforced composites are a desired material for many engineering applications where high strength is needed while maintaining low weight.  In order for composites to be used in structural components, efficient and accurate computer models need to be developed which can predict the mechanical response under different loading conditions.  Modeling composites can be difficult because a high fidelity is often needed to capture microscale features (such as groups of densely-packed fibers or resin-rich areas) which can be costly in computation time. To this point, multiscale models have been found to be beneficial in capturing microscale features in smaller representative models, where the mechanical response is homogenized to the macroscale \cite{stapleton2017multi,stapleton2018multi,pineda2022}. In its most basic form, the microscale can be modeled as a repeating unit cell (RUC) which consists of a few fibers uniformly arranged in a repeating pattern.  More recently, micorstrucutres are being modeled as representative volume elements (RVEs) which are theorized to be the minimum volume which can capture some bulk behavior.  RVEs typically have more fibers than RUCs, which results in more computation time, but are able to capture microscale features formed by fiber arrangements which may impact mechanical behavior \cite{kwon2008multiscale,qiao2021size,ghayoor2018micromechanical,mishnaevsky2014hybrid,li2015effects,bulsara1999damage,elnekhaily2018damage,sudhir2019simulation}. Generating RVEs of composite microstrucutres with statistically equivalent features as manufactured samples presents another challenge.  In the past, different statistics have been developed, such as Ripley's K Function, which are used to compare fiber distributions and determine equivalence, but these statistics often fall short of characterizing features such as fiber clusters or matrix pockets \cite{elnekhaily2018damage,ripley1977modelling}.  Furthermore, the generation of these microstrucutral models often rely on ``pick and place'' simulations where fibers are placed in a bounded area until a set value is reached or the simulation jams.  In these cases, it is difficult to control the fiber arrangement and create artificial microstrucutres with desired features while maintaining a degree of randomness \cite{feder1980random,hinrichsen1986geometry,buryachenko2003quantitative,yang2013new,park2019efficient}.

\begin{figure}[ht!]
	\centering
	\includegraphics[width=0.8\linewidth]{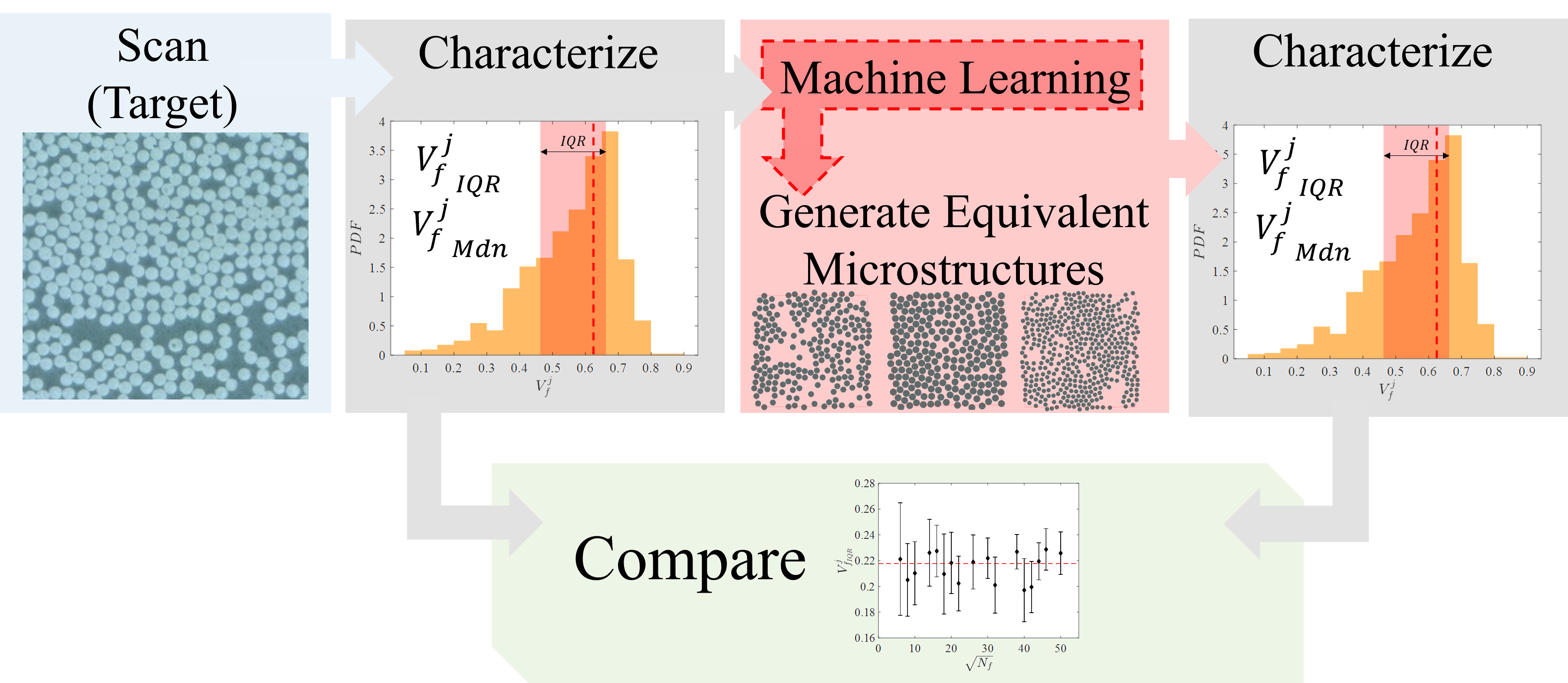}
	\caption{\label{fig:flowchart}
		Process of taking a characterized microstructure, or target characteristics, using the machine learning model to generate a set of equivalent microstructures, characterizing the generated equivalent set, and comparing the target features to what was generated.
	}
\end{figure}

In this study, an RVE generator based on the Discrete Element Method was used to randomly initialize an area with fibers and then apply contact between fibers during an explicit time stepping simulation.  Various inputs such as damping coefficients and seeding parameters were used to generate RVEs with features observed in experimentally-obtained micrographs.  A characterization method was used to describe a microstrucuture based on its local fiber volume fraction median and inter-quartile range (IQR), $V_{f}^{Mdn}$ and $V_{f}^{IQR}$ \cite{husseini2023generation,husseini2023generation2}.  These descriptors were used as targets or desired quantities of interest for the inverse problem for the RVE generator.  Equivalent microstrucutres were generated and characterized, and the results were compared to the original target value. A depiction of this process is shown in Figure \ref{fig:flowchart}

\subsection{Problem Statement}

 A random microstructure generator which uses input parameters to control the resulting fiber morphology while still maintaining a degree of randomness was utilized for this study \cite{husseini2023generation,husseini2023generation2}.  The generator was developed using in-house code with the Discrete Element Method as its foundation.  Input parameters of this generator were used to control the resulting fiber morphologies. Fiber seeding parameters control the size of the microstructure and the initial placement or seeding of the fibers prior to enforcing contact between fibers. First, a number of fibers, $n_f$, were placed into a bounded region divided into cells which were padded by a margin, $m_{cell}$, based on a user-defined number of fibers per cell, $n_{f/cell}$, until a global volume fraction, $V_f$, was reached.  Increasing the margin decreased the area where fibers could be placed, increasing the chance of initial overlapping and potential energy of the system. Once seeded, fibers were allowed to disperse, with movement initiated by contact between fibers. The simulation phase of this generator continued until a prescribed kinetic energy cutoff, $\epsilon_{KE}$, criteria was met.  Initially, a prescribed number of relaxation iterations was run without damping until a maximum number of time steps, $t_{n_{max}}$, was reached and damping was enforced. Three different types of damping were used to alter the kinetic energy dispersion: contact damping, $C_{ij}^{n}$, global damping, $C_i$,  which increased with each time step, and incremental damping, $d_i$, which was enforced at each time step. Once the simulation ended, a minimum spacing between the fibers, $d_{min}$, was created by reducing the radius of each fiber. For this study, $d_{min}$, $d_i$, $t_{n_{max}}$, and $\epsilon_{KE}$ were fixed while the machine learning model was used to identify the remaining inputs.

\begin{figure}[ht!]
	\centering
	\includegraphics[width=0.9\linewidth]{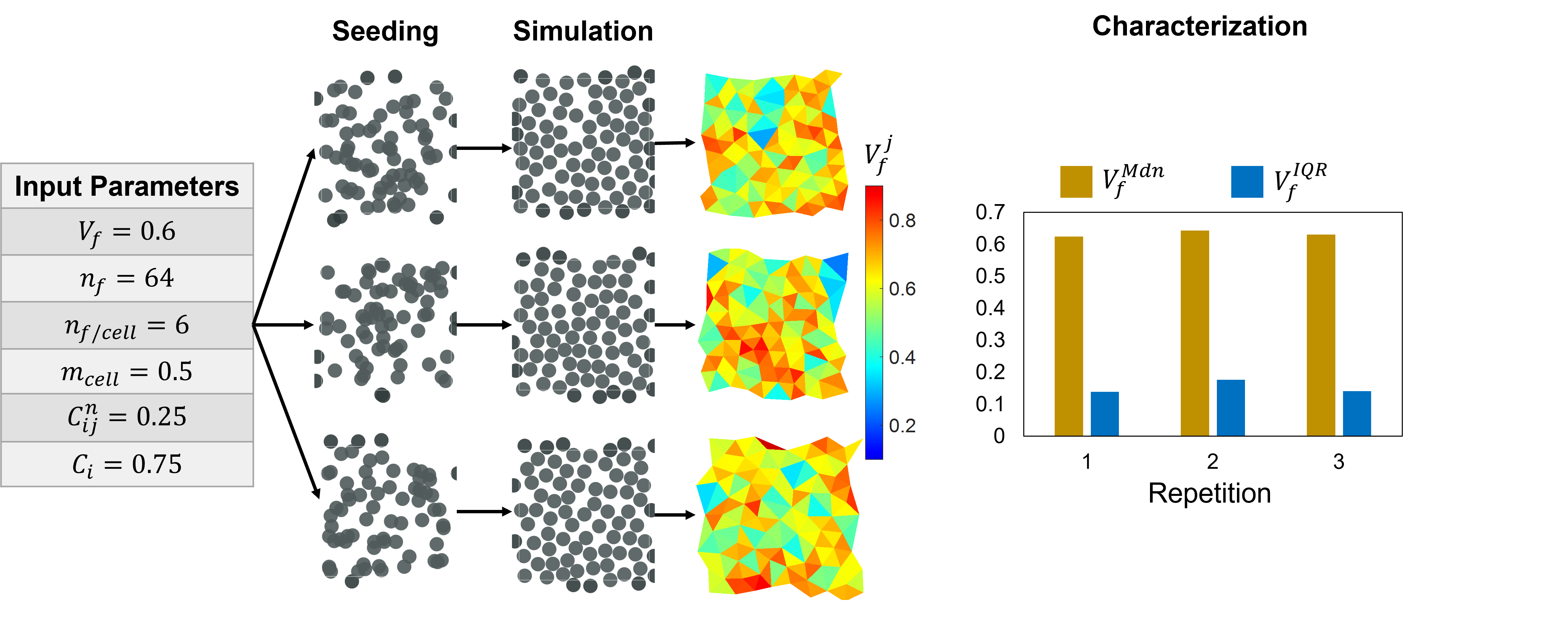}
	\caption{\label{fig:inputcombos}
		 One combination of input parameters produces an different seeding positions and therefore different microstructures and descriptors after characterization.   
	}
\end{figure}

One combination of inputs for the microstructure generator does not create a deterministic fiber morphology because of the initial random seeding of fibers as depicted in Figure \ref{fig:inputcombos}. The randomness in the generator comes from the random initial seeding where fibers are allowed to overlap.  Once the simulation starts, and depending on the initial seeding, the fibers disperse and collide until contact and viscous damping remove energy from the system and the fibers reach a relaxed state.  This randomness will result in different final fiber arrangements for different realizations, which is reflected by the fact that there are distributions of the descriptors for each input combination. Therefore, as explained in Section \ref{sec:prelim}, the $\delta y$ term in the observation model $y=f(\mathbf{x})+\delta y$ explains this inherent randomness. While the benefit of this generator remains high, the fact that one set of inputs does not yield a deterministic output adds additional complexity to the inverse problem.  To account for this, each set of inputs were repeated $50$ times to measure the average resulting microstructure descriptors used for training the machine learning model. For measuring the performance of the machine learning model solving the inverse problem, the same approach was taken where an input combination was repeated $50$ times to measure the average descriptors.  

\subsection{Results for Design Optimization with One Target Output}

\begin{figure}[ht!]
	\centering
	\includegraphics[width=0.9\linewidth]{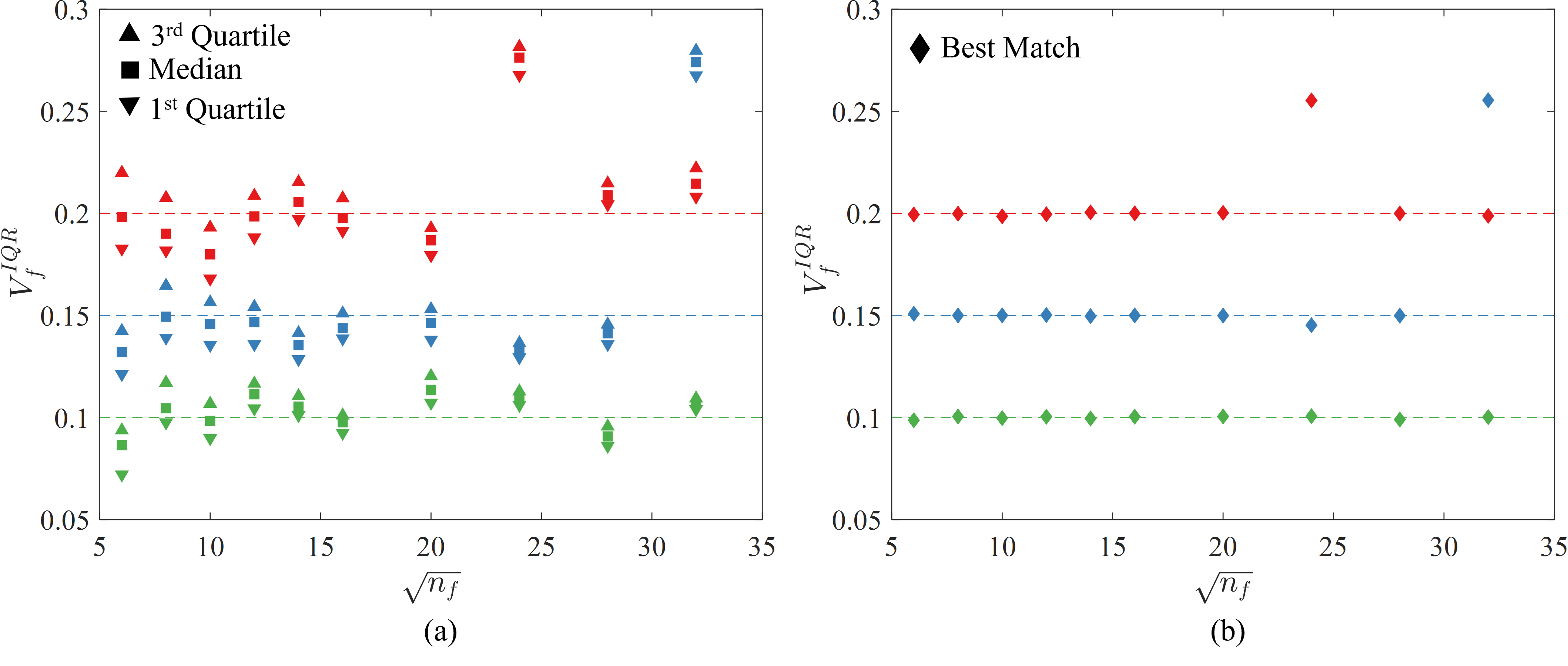}
	\caption{\label{fig:oldIQR}
		An experiment where a target $V_{f}^{IQR}$ was set at different numbers of fibers, $n_f$, using only the learner to provide inputs to the microstrucutre generator. a)  The median and IQR of each of the $50$ microstructures compared to their corresponding target, and b) the closest match to the target  from each batch of $50$.  
	}
\end{figure}

To see the improvement of the machine learning-based design optimization problem with both learner and evaluator steps, rather than just learner, two tests were conducted.  One target output for $V_{f}^{IQR}$ was defined for three different values and at different microstrucutre sizes by varying $n_f$. Target $V_{f}^{IQR}$ values were set at $0.1$, $0.15$, and $0.2$ for $n_f$ ranging from $36$ to $1,\!024$.  The first test used just the learner stage of the model to determine the rest of the generator inputs required to match each $V_{f}^{IQR}$ shown in Figure \ref{fig:oldIQR}. Throughout this section, we set $\alpha=0.1$ for conformal inference, $K=20$ for the number of folds in cross-validation, and the learner model is chosen to be a nearest neighbor regressor, where we perform a grid search to find the optimal number of neighbors. Specifically, the search space for this hyperparameter is defined as the set $\{2,5,10,15,20,25,30\}$. Note that we mainly focus on shallow machine learning models because of the low-dimensional structure of the input space. 

Figure \ref{fig:oldIQR}(a) shows the median, with the 1st and 3rd quartile of the $50$ generated microstrucutres shown for each target case.  Figure \ref{fig:oldIQR}(b) shows a single microstrucutre from each set which matched closest to the target and the performance is shown by how close a point lies to its corresponding dashed line.  These results show two cases where the prediction was far from the target and that there was not a microstrucuture in those sets that was close to the target. While for most of the targets there was at least one microstrucutre which matched closely, it can be seen how there may be outliers in the prediction.

\begin{figure}[ht!]
	\centering
	\includegraphics[width=0.9\linewidth]{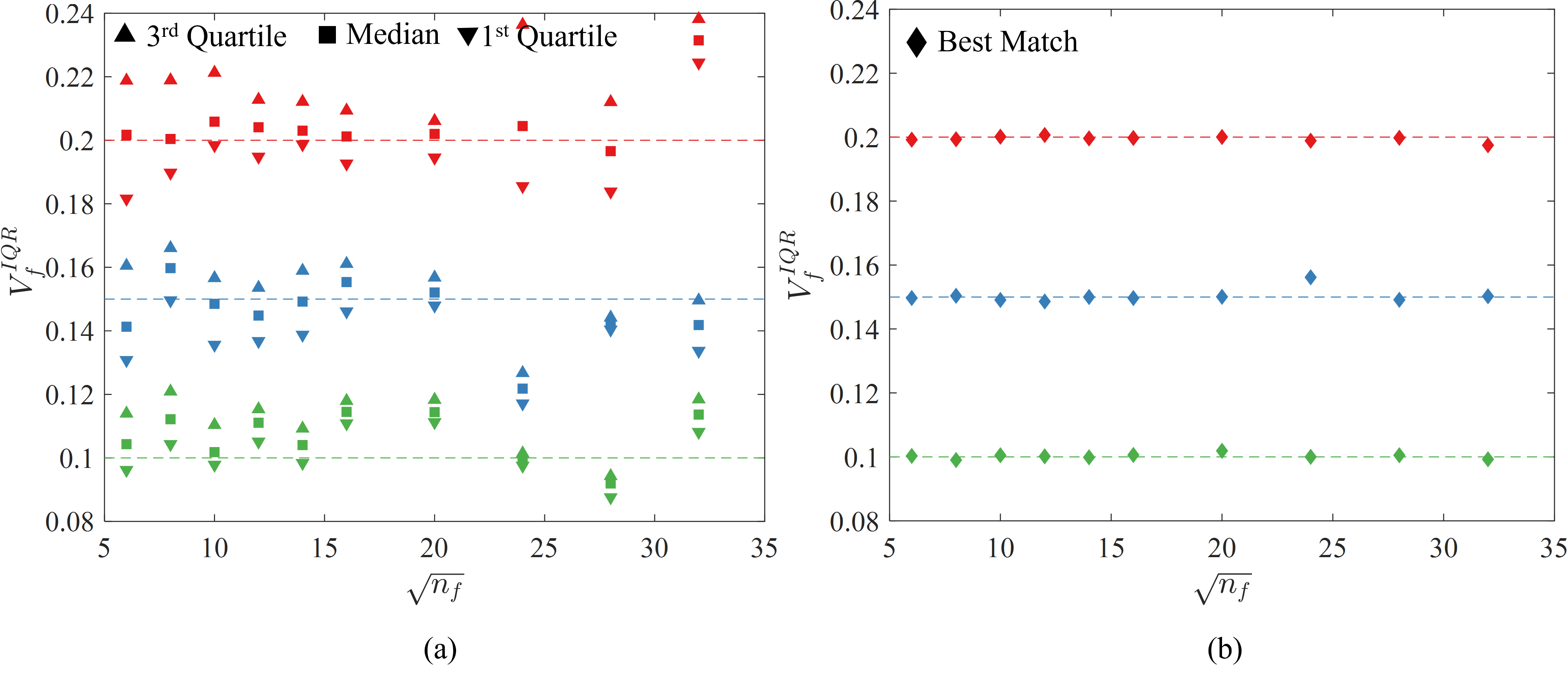}
	\caption{\label{fig:newIQR}
		An experiment where a target $V_{f}^{IQR}$ was set at different $n_f$ using the learner and evaluator to provide inputs to the microstrucutre generator. a)  Showing the median and IQR of each of the $50$ microstructures compared to their corresponding target, and b) the closest match to the target  from each batch of $50$.  
	}
\end{figure}

A case where the learner and evaluator model was used in our proposed two-stage framework is shown in Figure \ref{fig:newIQR}. The evaluator model is a polynomial regression model, and the degree of this polynomial model is selected from the set $\{2,3,4,5,6\}$. The same targets were used in Figure \ref{fig:newIQR} and Figure \ref{fig:oldIQR}.  It can be seen how the two outliers in the case with just the learner are no longer present in the case with the learner and evaluator.  Two new inputs were found by the evaluator which ended up matching much closer and producing at least one microstrucutre which matched the target well.  

Therefore, the proposed framework excels in identifying input conditions or configurations that closely align with the desired outputs at a reasonable level, employing relatively simple machine learning models such as the nearest neighbor and polynomial regressors. It is important to note that the primary computational cost of our framework is associated with the memory and time requirements linked to training these machine learning models. While opting for more sophisticated machine learning models, such as neural networks, may enhance performance, a practical consideration involves the available budget for design optimization. Similar considerations apply to the use of advanced hyperparameter optimization techniques, such as Bayesian optimization, which are typically more resource-intensive than simpler methods like random or grid search. However, our proposed framework incorporates safety measures against inaccurate and uncertain solutions. Therefore, the prioritization of cost-effective machine learning models is recommended unless the conformal prediction procedure considers all candidate solutions to be ineligible.

\begin{figure}[ht!]
	\centering
	\includegraphics[width=0.9\linewidth]{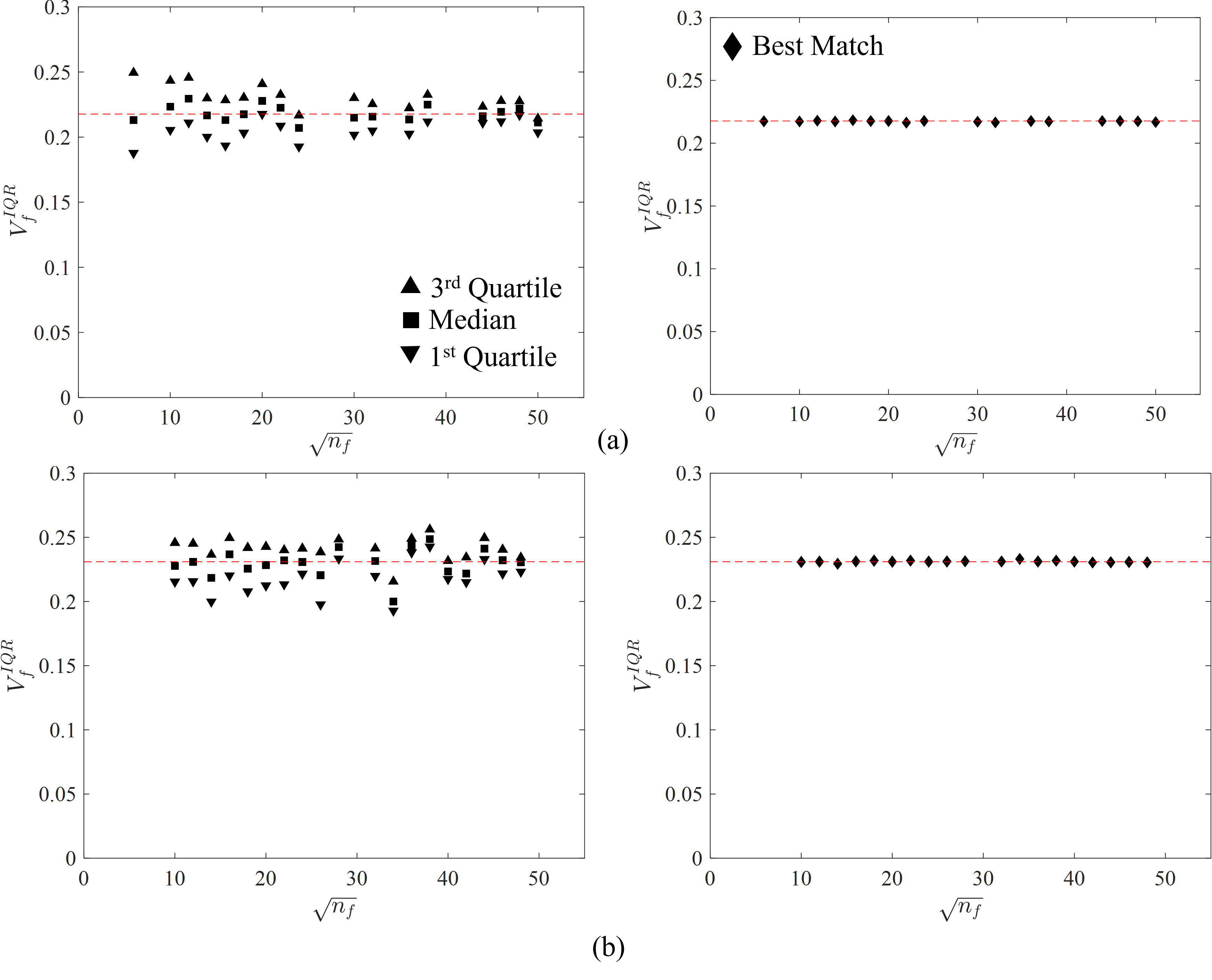}
	\caption{\label{fig:scans}
		An experiment where the $V_{f}^{IQR}$ was measured from two real microstructure scans a) and b) and used as the target for varied $n_f$.  The median and IQR of the results for creating equivalent microstrucutres to each scan are shown as well as the best match from each batch from the same input.  
	}
\end{figure}

To show a real application, two  microstructures obtained for manufactured composites were characterized and the measured $V_{f}^{IQR}$s were used as a targets for different $n_f$ in Figure \ref{fig:scans}. Figure \ref{fig:scans}(a) and Figure \ref{fig:scans}(b) show similar results that for each  $n_f$, the machine learning model was able to produce generator inputs which would yield artificial microstructures with equivalent $V_{f}^{IQR}$ values. The median and quartiles fall very close to the target for both cases meaning that a majority of the $50$ generated microstructures were close to the target for a given input.  Additionally, there is at least one artificial microstrucutre for each $n_f$ which matches very closely to the scan.  There are cases where there is no data; this is because the evaluator could not find a good solution.

\subsection{Results for Design Optimization with Two Target Outputs}

\begin{figure}[ht!]
	\centering
	\includegraphics[width=0.9\linewidth]{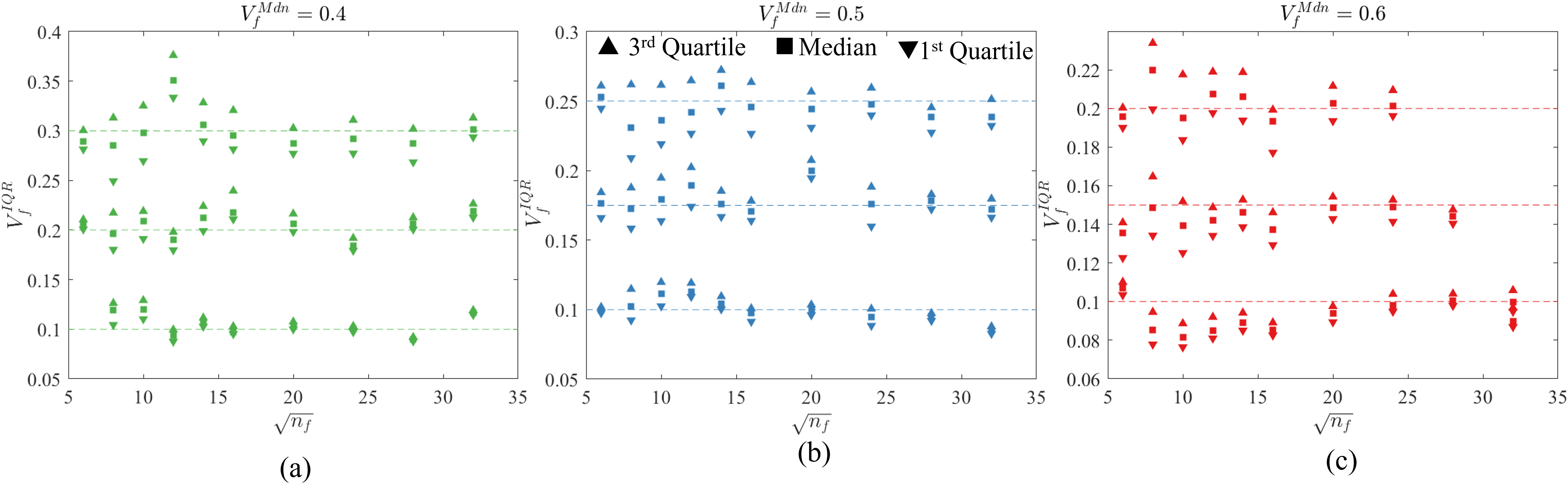}
	\caption{\label{fig:twotarget}
		An experiment where a target $V_{f}^{IQR}$ and $V_{f}^{Mdn}$ was set at different values of $n_f$ using the learner and evaluator to provide inputs to the microstrucutre generator. a)  The median and IQR of each of the $50$ microstructures are shown compared to their corresponding target, and b) the closest match to the target  from each batch of $50$.
	}
\end{figure}

An experiment was run where two targets were set, $V_{f}^{IQR}$ and $V_{f}^{Mdn}$.  $V_{f}^{Mdn}$ was set at $0.4$, $0.5$, and $0.6$. For each $V_{f}^{IQR}$, three $V_{f}^{IQR}$s were set, which varied depending on $V_{f}^{Mdn}$.  Physically, microstrucutres with a higher $V_{f}^{Mdn}$ could not achieve a high  $V_{f}^{IQR}$ because at high $V_{f}^{Mdn}$ the fibers are approaching their highest-packed state.  This means that there is less variability in the arrangement and the local fiber volume fraction distribution becomes more narrow, decreasing $V_{f}^{IQR}$.  At each target, $50$ inputs were generated for varied $n_f$, and compared in Figure \ref{fig:twotarget}.

Figure \ref{fig:twotarget} shows similar trends to the single target experiments.  Typically, the median of the $50$ generated microstrucutres fell very closely to the targets while the IQR varied.  It can also be seen in Figure \ref{fig:twotarget}(c) that there are some data points missing, meaning that there were no confident solutions for those targets.  The results of this study show that for a single or multiple targets, the two-stage framework can produce inputs to the microstrucutre generator which will yield at least one microstrucutre that matches the target.

\section{Conclusion and Future Work}\label{sec:conc}
In conclusion, this paper has presented a two-stage surrogate modeling framework that significantly enhances the accuracy and reliability of design optimization problems. The key novelty lies in the successful integration of conformal inference, enabling seamless interactions between two machine learning surrogates. One surrogate acts as a forward problem proxy, identifying promising solutions, while the other plays an advisory role, effectively eliminating inaccurate or uncertain outcomes. The advantages of the introduced framework over traditional single-stage inverse problems are twofold. Firstly, it obviates the need for intensive hyperparameter optimization, as the evaluator model filters out undesired solutions in the second stage, as demonstrated through comprehensive experiments. Secondly, the proposed framework exhibits remarkable applicability across a wide range of problems, owing to conformal inference's minimal distributional and model assumptions. By integrating conformal inference with any standard regression model, the framework readily provides prediction intervals, making it versatile and readily adaptable. Our experiments revealed that commonly selected values for $\alpha$ in conformal prediction, such as 0.1 or 0.2, provide safety measures against inaccurate solutions obtained by the learner model. A comparison between Figures \ref{fig:oldIQR} and \ref{fig:newIQR} highlighted that the implementation of the two-stage surrogate modeling framework results in microstructures with descriptors closely situated within a $0.01$ neighborhood of the target output. Furthermore, we demonstrated the efficacy of our framework in addressing design optimization problems with multiple quantities of interest.

In light of the versatile framework presented in this work for solving design optimization problems, there exist several promising avenues for future research that can enhance interactions between machine learning surrogates. One crucial direction is to explore the performance of conformal inference and the proposed framework when dealing with extreme or rare events within regression problems. Conducting an in-depth analysis by adjusting the parameters of the studied Ishigami function to generate outputs with varying skewness levels will provide valuable insights. Understanding how the framework behaves under such challenging scenarios can lead to the development of robust techniques to handle extreme events and improve the reliability of predictions in real-world applications. Another impactful research direction is to design innovative protocols that allow more than two machine learning surrogates to interact collaboratively, thereby offering the potential to remove even more undesired solutions from inverse problems. Analyzing the tradeoffs between computational efficiency and accuracy improvements will be crucial in evaluating the feasibility and effectiveness of such protocols. This investigation could lead to the development of sophisticated ensemble techniques or decision-making strategies, enabling the framework to leverage multiple surrogates' insights and further enhance its performance. Moreover, an essential aspect of future research would involve exploring the framework's scalability to handle large-scale and high-dimensional optimization problems.

\section*{Acknowledgements}
This work was partially supported by a NASA Space Technology Graduate Research Opportunity (80NSSC21K1285) and by a NASA NRA (80NSSC21N0102) through the NASA Transformational Tools and Technologies (TTT) program, under the Aeronautics Research Mission Directorate (ARMD). 
This work was completed in part with resources provided by the University of Massachusetts' Green High Performance Computing Cluster (GHPCC).

\bibliographystyle{ieeetr}
\bibliography{sample}

\end{document}